\relax
\documentclass[letterpaper]{article} 
\usepackage{proceed}
\usepackage[margin=1in]{geometry}
\usepackage{times}  
\usepackage{url}  
\usepackage{graphicx}  
\usepackage{xcolor} 
\usepackage{makecell}
\usepackage{caption}
\usepackage{subcaption}
\usepackage{amsmath}
\usepackage{dsfont}
\usepackage{breqn}
\usepackage{xr}
\usepackage{amssymb}
\usepackage{apacite}
\usepackage{tabularx}
\usepackage{booktabs}
\usepackage{xr}

\definecolor{Green}{RGB}{58, 147, 48}

\newcommand{\mcrl}{metacognitive reinforcement learning}

\makeatletter
\newcommand*{\addFileDependency}[1]{UAI2018.tex
  \typeout{(#1)}
  \@addtofilelist{#1}
  \IfFileExists{#1}{}{\typeout{No file #1.}}
}
\makeatother

\begin{document}

\title{Learning to select computations}
\author{Frederick Callaway\textsuperscript{1,a}, Sayan Gul\textsuperscript{1,a}, Paul Krueger\textsuperscript{a}, Thomas L. Griffiths\textsuperscript{a}, \& Falk Lieder\textsuperscript{1,a,b} \\
\textsuperscript{a} University of California, Berkeley, USA\\
\textsuperscript{b} Max Planck Institute for Intelligent Systems, T\"ubingen, Germany\\
\textsuperscript{1} These authors contributed equally.
}
\maketitle
\begin{abstract}

The efficient use of limited computational resources is an essential ingredient of intelligence. Selecting computations optimally according to rational metareasoning would achieve this, but this is computationally intractable.
Inspired by psychology and neuroscience, we propose the first concrete and domain-general learning algorithm for approximating the optimal selection of computations: Bayesian metalevel policy search (BMPS). We derive this general, sample-efficient search algorithm for a computation-selecting metalevel policy based on the insight that the value of information lies between the myopic value of information and the value of perfect information.
We evaluate BMPS on three increasingly difficult metareasoning problems: when to terminate computation, how to allocate computation between competing options, and planning. Across all three domains, BMPS achieved near-optimal performance and compared favorably to previously proposed metareasoning heuristics. Finally, we demonstrate the practical utility of BMPS in an emergency management scenario, even accounting for the overhead of metareasoning.
\end{abstract}

\section{1\space\space\space\space INTRODUCTION}

The human brain is the best example of an intelligent system we have so far. One feature that sets it apart from current AI is the remarkable computational efficiency that enables people to effortlessly solve hard problems for which artificial intelligence either under-performs humans or requires superhuman computing power and training time.
For instance, to defeat Garry Kasparov $3.5$--$2.5$, Deep Blue had to evaluate $200\,000\,000$ positions per second, whereas Kasparov was able to perform at almost the same level by evaluating only $3$ positions per second \cite{DeepBlue,IBM}. This ability to make efficient use of limited computational resources is the essence of intelligence \cite{RussellWefald1991}.
People accomplish this feat by being very selective about when to think and what to think about, choosing computations adaptively and terminating deliberation when its expected benefit falls below its cost \cite{Gershman2015,LiederGriffiths2017,Payne1988}.

Rational metareasoning was introduced to recreate such intelligent control over computation in machines \cite{Horvitz1989,Russell1991,Hay2012}. In principle, rational metareasoning can be used to always select those computations that make optimal use of the agent's finite computational resources. However, its computational complexity is prohibitive \cite{Hay2012}. The human mind circumvents this computational challenge by learning to select computations through metacognitive reinforcement learning \cite{KruegerLieder2017,LiederGriffiths2017,Wang2017}. Concretely, people appear to learn to predict the value of alternative cognitive operations from features of the task, their current belief state, and the cognitive operations themselves. If humans learn to metareason through \mcrl, then it should be possible to build intelligent systems that learn to metareason as efficiently as people.

In this paper, we introduce Bayesian metalevel policy search (BMPS), the first domain-general algorithm for learning how to metareason, and evaluate it against existing methods for approximate metareasoning on three increasingly more complex toy problems. Finally, we show that our method makes metareasoning efficient enough to offset its cost in a more realistic emergency management scenario. In this problem, which we use as a running example, an emergency manager must decide which cities to evacuate in the face of an approaching tornado. She bases her decision on a series of computationally intensive simulations that noisily estimate the impact of the tornado on each city. Because time is short, she is forced to decide which simulations are the most important to run. In the following section, we discuss how to formalize this problem as a sequential decision process.

\newcommand{\A}{\mathcal{A}}
\newcommand{\B}{\mathcal{B}}
\newcommand{\C}{\mathcal{C}}
\newcommand{\meta}{_{\text{meta}}}
\newcommand{\Qmeta}{$Q\meta ^\star$}
\newcommand{\expect}[2][]{\mathds{E}_{#1} \left[ #2 \right]}
\newcommand{\state}{\theta}
\newcommand{\uthetapi}{U_\pi(\theta)}
\newcommand{\ubpi}{\hat{U}_\pi(b)}
\newcommand{\Etheta}[1]{\mathbb{E}_{\theta \sim b} [ #1 ]}
\newcommand{\Vpi}{V_\pi^{(\theta)}(s)}

\section{2\space\space\space\space BACKGROUND}
\subsection{2.1\space\space\space\space METAREASONING}

If reasoning seeks an answer to the question ``what should I do?'', metareasoning seeks to answer the question ``how should I decide what to do?''. 
The theory of rational metareasoning \cite{Russell1991,Russell1995} frames this problem as selecting computations so as to maximize the sum of the rewards of resulting decisions minus the costs of the computations involved. Concretely, one can formalize reasoning as a metalevel Markov decision process (metalevel MDP) and metareasoning as solving that MDP \cite{Hay2012}.
While traditional (object-level) MDPs describe the objects of reasoning---the state of the external environment and how it is affected by physical actions---a metalevel MDP describes reasoning itself. Formally, a metalevel MDP
$
M\meta = \left(\B, \A, T\meta, r\meta \right)
$
is an MDP where the states $\B$ encode the agent's beliefs, the actions $\A$ are computations, the transition function $T\meta$ describes how computations update beliefs, and the reward function $r\meta$ describes the costs and benefits of computation. A definition table for our notation is included in the Supplementary Material.

A belief state $b \in \B$ encodes a probability distribution over parameters $\state$ of a model of the domain. 
For example, in the tornado problem described in the introduction, $\state$ could be a vector of $k$ probabilities that each of the $k$ cities will incur evacuation-warranting damage; $b$ would thus encode $k$ distributions over $[0, 1]$, e.g. $k$ Beta distributions.
The parameters $\state$ determine the utility of acting according to a policy $\pi$, that is $\uthetapi$. For one-shot decisions, $\uthetapi$ is the expected reward of taking the single action identified with $\pi$. In the tornado problem, for example, $\pi$ can be represented as a binary vector of length $k$ indicating whether each city should be evacuated, and $\uthetapi$ is the cost of making the evacuations plus the expected cost of failing to evacuate cities that incur major damage.
In sequential decision-problems, $\uthetapi = \Vpi$ is the expected sum of rewards the agent will obtain by acting according to policy $\pi$ if the environment has the characteristics encoded by $\state$.

$\A$ includes computations $\C$ that update the belief, as well as a special metalevel action $\bot$ that terminates deliberation and initiates acting on the current belief. The effects of computations are encoded by $T\meta: \B \times \A \times \B \rightarrow [0,1]$ analogously to a standard transition function. The termination action always leads to a unique end state.

The metalevel reward function $r\meta$ captures the cost of thinking \cite{Shugan1980} and the external reward the agent expects to receive from the environment. The computations $\C$ have no external effects and thus always incur a negative reward  $r\meta(b,c)=-\text{cost}(c)$. In the problems studied below, all computations that deliberate have the same cost, that is $\text{cost}(c)=\lambda$ for all $c \in \C$ whereas $\text{cost}(\bot)=0$. An external reward is received only when the agent terminates deliberation and makes a decision, which is assumed to be optimal given the current belief. The metalevel reward for terminating is thus $r\meta(b,\bot) = \max_\pi \Etheta{\uthetapi}$.%
\footnote{If the agent's model is unbiased, this reward has the same expectation but lower variance than the true external reward.}

Early work on rational metareasoning \cite{Russell1991} defined the optimal way to select computations as maximizing the value of computation (VOC):
\begin{equation}
    \pi^*\meta = \arg\max_c \text{VOC}(c,b),\label{eq:optimalPolicy}
\end{equation}
where $\text{VOC}(c,b)$ is the expected improvement in decision quality that can be achieved by performing computation $c$ in belief state $b$ and continuing optimally, minus the cost of the optimal sequence of computations \cite{Russell1991}. When no computation has positive value, the policy terminates computation and executes the best object-level action, thus $\text{VOC}(\bot,b) = 0$.

\subsection{2.2\space\space\space\space APPROXIMATE METAREASONING}

Previous work \cite{Russell1991,Lin2015} has approximated rational metareasoning by the meta-greedy policy
$\arg\max_c \text{VOC}_1(c,b)$
where
$
    \text{VOC}_1(c,b) = \expect[B'\sim T\meta(b,c,\cdot)]{r\meta(B', \bot)} - r\meta(b,\bot) + r\meta(b,c),
$
is the myopic value of computation \cite{Russell1991}. The meta-greedy policy selects each computation assuming that it will be the last computation. 
This policy is optimal when computation provides diminishing returns (i.e. the improvement from each additional computation is less than that from the previous one), but it deliberates too little when this assumption is violated.
For example, in the tornado problem (where false negatives have high cost), a single simulation may be unable to ensure that evacuation is unnecessary with sufficient confidence, while two or more could.

Hay et al. (2012) approximated rational metareasoning by combining the solutions to smaller metalevel MDPs that formalize the problem of deciding how to decide between one object-level action and the expected return of its best alternative.
Each of these smaller metalevel MDPs includes only the computations for reasoning about the expected return of the corresponding object-level action.
While this \textit{blinkered} approximation is more accurate than the meta-greedy policy, it is also significantly less scalable and not directly applicable to metareasoning about planning. 

These are the main approximations to rational metareasoning. So, to date, there appears to be no accurate and scalable method for solving general metalevel MDPs.

\subsection{2.3\space\space\space\space METACOGNITIVE RL}

It has been proposed that metareasoning can be made tractable by learning an approximation to the value of computation \cite{Russell1991}. However, despite some preliminary steps in this direction \cite{Harada1998,lieder2014algorithm,LiederKrueger2017} and related work on meta-learning \cite{Smith-Miles2009,Thornton2013,Wang2017}, learning to approximate bounded optimal information processing remains an unsolved problem in artificial intelligence.

Previous research in cognitive science suggests that people circumvent the intractability of metareasoning by learning a metalevel policy from experience \cite{LiederGriffiths2017,Cushman2015,KruegerLieder2017}. At least in some cases, the underlying mechanism appears to be model-free reinforcement learning (RL) \cite{Cushman2015,KruegerLieder2017}.
This suggests that model-free reinforcement learning might be a promising approach to solving metalevel MDPs. To our knowledge, this approach is yet to be explored in artificial intelligence. Here, we present a proof-of-concept that near-optimal metalevel policies can be learned through metacognitive reinforcement learning.

\begin{figure}[hb!]
  \centering
  \includegraphics[width=0.45\textwidth]{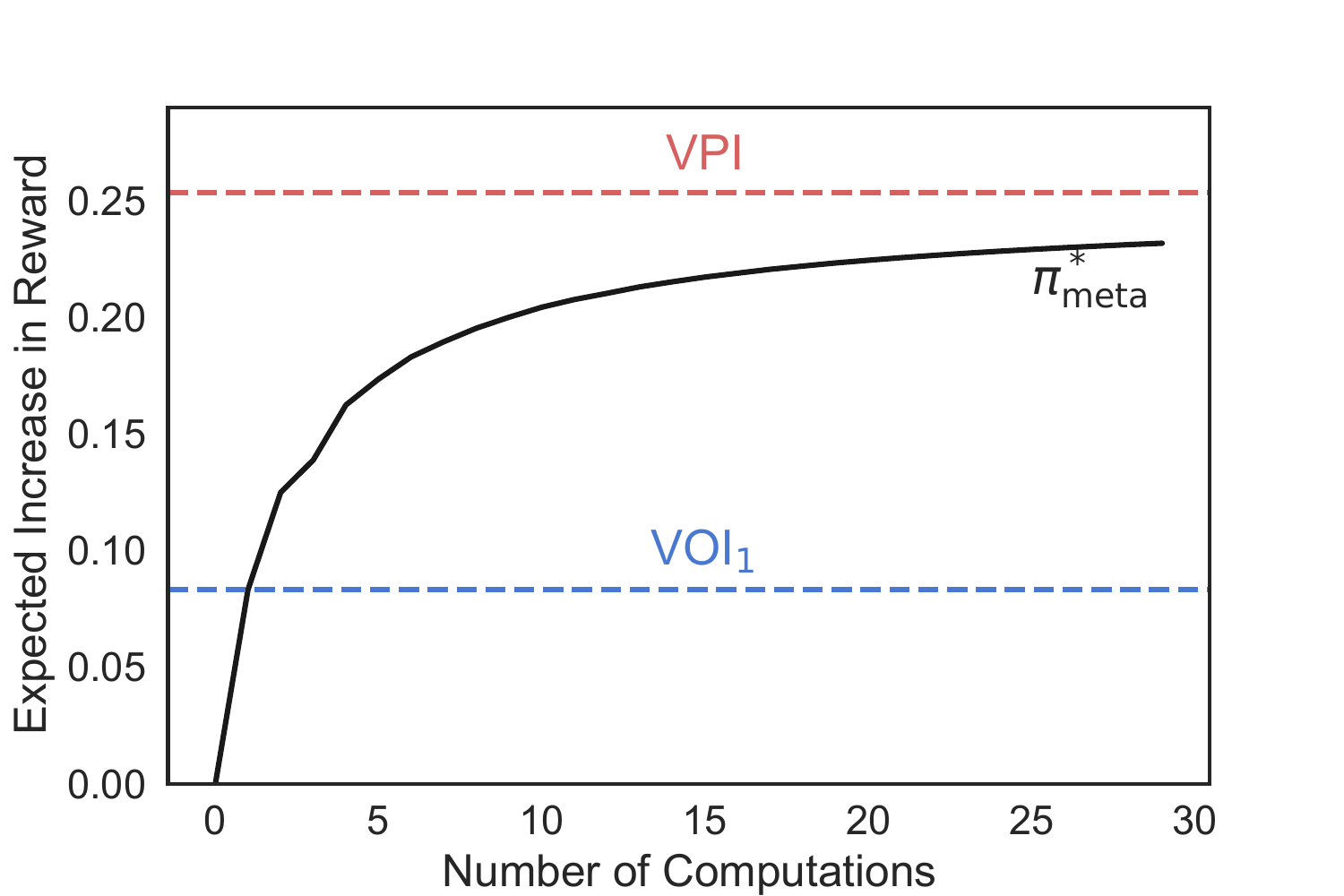}
  \caption{Expected performance in metareasoning about how to choose between three actions increases monotonically with the number of computations, asymptoting at the value of perfect information (VPI). Consequently, the value of executing a single computation must lie between the myopic value of information ($\text{VOI}_1$) and the VPI.}
  \label{fig:VOC}
 \end{figure}

\section{3\space\space\space\space BAYESIAN METALEVEL POLICY SEARCH}
\newcommand{\VPIall}{\text{VPI}}
\newcommand{\VPIsub}{\text{VPI}_{\text{sub}}}
According to rational metareasoning, an optimal metalevel policy is one that maximizes the VOC (Equation~\ref{eq:optimalPolicy}). Although the VOC is intractable to compute, it can bounded. Bayesian metalevel policy search (BMPS) capitalizes on these bounds to dramatically reduce the difficulty of learning near-optimal metalevel policies.
Figure \ref{fig:VOC} illustrates that if the expected decision quality improves monotonically with the number of computations, then the improvement achieved by the optimal sequence of computations should lie between the benefit of deciding immediately after the first computation and the benefit of obtaining perfect information \cite{Howard1966}. The former is given by the myopic  value of information,\footnote{The $\text{VOI}_1$ defined here is equal to the myopic VOC defined by Russell and Wefald (1991) plus the cost of the computation.}
%
\begin{dmath}\label{eq:VOI1}
\text{VOI}_1(c, b) = \mathds{E}_{B'\sim T\meta(b,c,\cdot)} \left[ 
    U \left( B' \right) 
\right] - U \left( b \right)
.
\end{dmath}

and the latter is given by the value of perfect information,
\begin{dmath}\label{eq:VPIall}
\VPIall(b) = \expect[\theta^* \sim b]{
  U \left( B^* \left( \cdot; \; \theta^* \right) \right)
} - U \left( b \right)
,
\end{dmath}
where $U(b) = r\meta(b, \bot)$ is shorthand for the expected value of terminating computation and $B^*(\theta; \theta^*) = \delta(\theta_i - \theta^*_i)$ is the belief state with perfect knowledge of the true environment parameters $\theta^*$.

In problems with many parameters, this upper bound can be very loose because the optimal metalevel policy might reason only about a small subset of relevant parameters. 
To capture this, we introduce an additional feature $\VPIsub(c, b)$ that measures how beneficial it would be to have full information about a subset of the parameters that are most relevant to the given computation. We model relevance with a function $f(c,i)$ that returns $1$ if $\theta_i$ is relevant to what $c$ is reasoning about and $0$ otherwise. Using this relevance function, we define the value of gaining perfect information about the relevant subset of parameters as
%
%
\begin{dmath}\label{eq:VPIsub}
\VPIsub(c, b) = \expect[\theta^* \sim b]{
  U(B'_{\text{sub}}(\cdot; c, b, \theta^*))
} - U(b)
,   
\end{dmath} 
with
%
$$
B'_{\text{sub}}(\theta; \; c, b, \theta^*) = \prod_i^k
  B^*(\theta_i; \theta^*)^{f(c,i)} \cdot
  b(\theta_i)^{1-f(c,i)},
$$
where $k$ is the number of parameters in the agent's model of the environment.
In the tornado problem, for example, each simulation is informative about a single parameter (the probability that the target city will sustain evacuation-warranting damage); thus, we define $f(c_j, i) = \mathds{1}(j = i)$. In the general case, the relevance function is a design choice that affords an easy opportunity to imbue BMPS with domain knowledge. In the simulations reported below, the relevance function associates each $c$ with the set of parameters that inform the value of the actions (or, in the case of planning, options) that $c$ reasons about.

Critically, all three VOI features can be computed efficiently or can be efficiently approximated by Monte-Carlo integration \cite{Hammersley2013}. BMPS thus approximates the VOC by a mixture of VOI features and an estimate of the cost of future computations
\newcommand{\VOCapprox}{\hat{\text{VOC}}(c,b; \mathbf{w})}
\begin{dmath}\label{eq:VOCApproximation}
    \VOCapprox = w_1 \cdot \text{VOI}_1(c,b) + w_2 \cdot \VPIall(b) + w_3 \cdot \VPIsub(c,b) - w_4\cdot \text{cost}(c),
\end{dmath}
with the constraints that $w_1,w_2,w_3 \in [0,1]$, $w_1+w_2+w_3=1$, and $w_4 \in [1,h]$ where $h$ is an upper bound on how many computations can be performed.
Since the VOC defines the optimal metalevel policy (Equation~\ref{eq:optimalPolicy}), we can define an approximately optimal policy,
$\pi\meta(b;\mathbf{w}) = \arg\max_c \VOCapprox$.
%

The parameters $\mathbf{w}$ of this policy are optimized by maximizing the expected return
$\mathds{E}\left[ \sum_t r\meta(b_t,\pi\meta(b_t; \mathbf{w})) \right]$, i.e. direct policy search. Because there are only three free parameters with the summation constraint, we propose using Bayesian optimization (BO) \cite{Mockus2012} to optimize the weights in a sample efficient manner.

The novelty of BMPS lies in leveraging machine learning to approximate the solution to metalevel MDPs and in the discovery of features that make this tractable. As far as we know, BMPS is the first general approach to metacognitive RL.
In the following sections, we validate the assumptions of BMPS, evaluate its performance on increasingly complex metareasoning problems, compare it to existing methods, and discuss potential applications.

\section{4\space\space\space\space EVALUATIONS OF BMPS}
We evaluate how accurately BMPS can approximate rational metareasoning against two state-of-the-art approximations---the meta-greedy policy and the blinkered approximation---on three increasingly difficult metareasoning problems.

\subsection{4.1\space\space\space\space WHEN TO STOP DELIBERATING?} 
How long should an agent deliberate before answering a question? Our evaluation mimics this problem for a binary prediction task (e.g., ``Will the price of the stock go up or down?''). Every deliberation incurs a cost and provides probabilistic evidence $X_t \sim \text{Bernoulli}(\theta)$ in favor of one outcome or the other. 
At any point the agent can stop deliberating and predict the outcome supported by previous deliberations.
The agent receives a reward of $+1$ if its prediction is correct, or incurs a loss of $-1$ if it is incorrect. The goal is to maximize the expected reward of this one prediction minus the cost of computation.

\subsubsection{4.1.1\space\space\space\space Metalevel MDP}
We formalize the problem of deciding when to stop thinking as a metalevel MDP $M\meta=(\B,\A,T\meta,r\meta)$ where each belief state $\left(\alpha,\beta\right) \in \B$ defines a beta distribution over the probability $\theta$ of the first outcome. 
The metalevel actions $\A$ are $\lbrace c_1, \bot \rbrace$ where $c_1$ refines the belief by sampling, and $\bot$ terminates deliberation and predicts the outcome that is most likely according to the current belief. The transition probabilities for sampling are defined by the agent's belief state, that is
$T\meta((\alpha,\beta),c_1,(\alpha+1,\beta))=\frac{\alpha}{\alpha+\beta}$ and $T\meta((\alpha,\beta),c_1,(\alpha,\beta+1))=\frac{\beta}{\alpha+\beta}$. 
The reward function $r\meta$ reflects the cost of computation, $r\meta(b,c_1)=-\lambda$, and the probability of making the correct prediction, $r\meta(b,\bot)=+1 \cdot p_{\text{correct}}(\alpha,\beta) - 1\cdot (1-p_{\text{correct}}(\alpha,\beta))$, where $p_{\text{correct}}(\alpha,\beta)=\max\lbrace \frac{\alpha}{\alpha+\beta}, \frac{\beta}{\alpha+\beta} \rbrace$). We set the horizon to $h=30$, meaning that the agent can perform at most $29$ computations before making a prediction (the $30$th metalevel action must be $\bot$).

Since there is only one parameter ($\theta$ has length one), the $\VPIsub$ feature is identical with the $\VPIall$ feature; thus, we exclude it. For the same reason, the blinkered approximation is equivalent to solving the problem exactly, and we exclude it from the comparison.

\subsubsection{4.1.2\space\space\space\space Evaluation procedure}
We evaluated the potential of BMPS in two steps: First, we performed a regression analysis to evaluate whether the proposed features are sufficient to capture the value of computation, computed exactly by backward induction \cite{Puterman2014}. Second, we tested whether a near-optimal metalevel policy can be learned by Bayesian optimization of the weights of the metalevel policy. We ran $500$ iterations of optimization, estimating the expected return of the policy entailed by the probed weight vector by its average return across $2500$ episodes. The performance of the learned policy was evaluated on an independent test set of $3000$ episodes.


\subsubsection{4.1.3\space\space\space\space Results} 
First, linear regression analyses confirmed that the three features ($\text{VOI}_1(c,b)$, $\text{VPI}(c,b)$, and $\text{cost}(c)$) are sufficient to capture between $90.8\%$ and $100.0\%$ of the variance in the value of computation for performing a simulation ($\text{VOC}(b,c_1)$) across different states $b$, depending on the cost of computation.

Concretely, as the cost of computation increased from $0.001$ to $0.1$ the regression weights shifted from $0.76\cdot \text{VPI} + 0.46 \cdot \text{VOI}_1 - 4.5 \cdot \text{cost}$ to $0.00\cdot \text{VPI} + 1.00 \cdot \text{VOI}_1 - 1.00 \cdot \text{cost}$ and the explained variance increased from $90.8\%$ to $100.0\%$. The explained variance and the weights remained the same for costs greater than $0.1$. Supplementary Figure 1 illustrates this fit for $\lambda=0.02$.

\begin{figure}
   \includegraphics[width=0.45\textwidth]{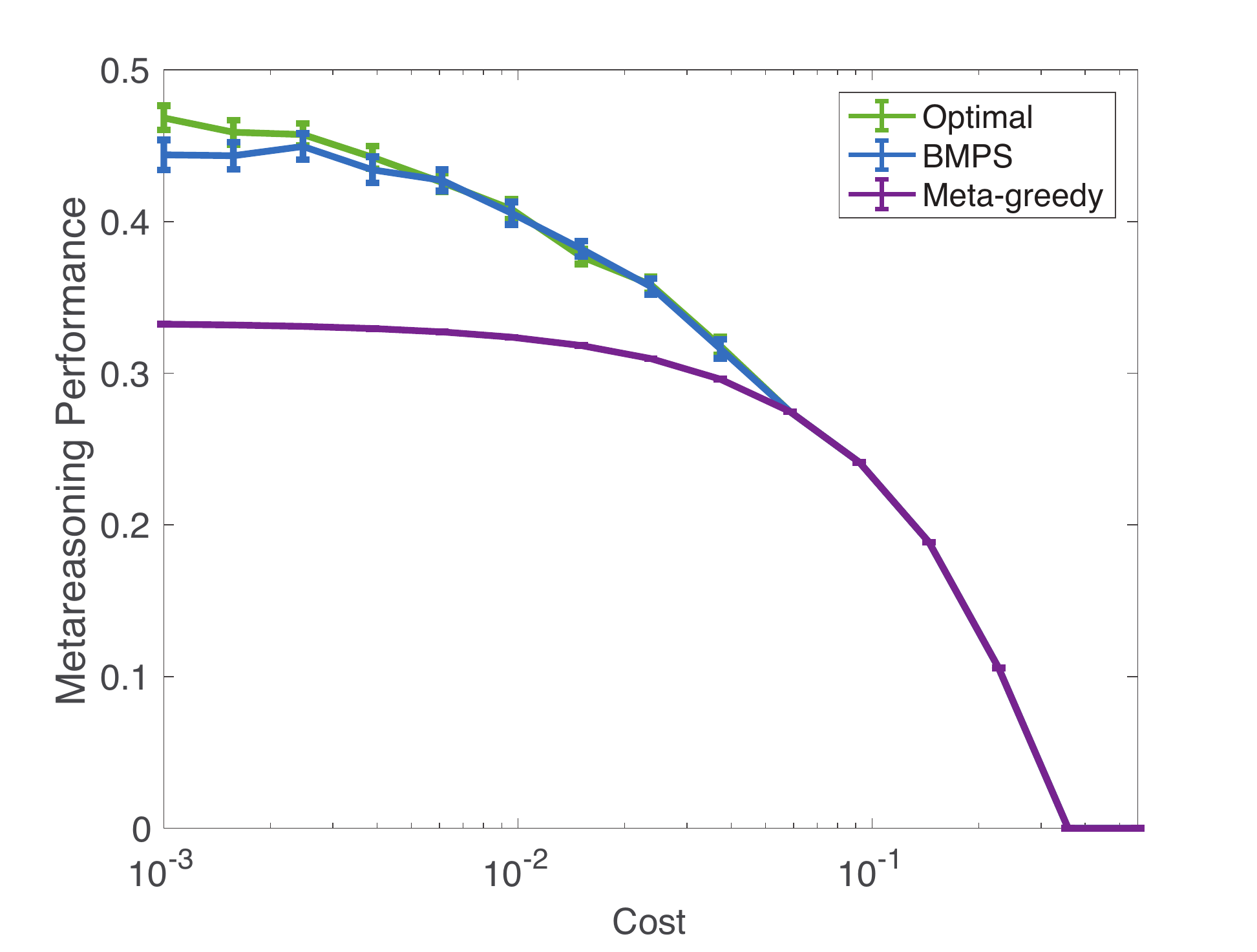}
   \caption{Results of performance evaluation on the problem of metareasoning about when to stop deliberating. }
   \label{fig:1LBPerformance}
 \end{figure}

Second, we found that the $\text{VOI}_1$ and the $\text{VPI}$ features are sufficient to learn a near-optimal metalevel policy. 
As shown in Figure \ref{fig:1LBPerformance}, the performance of BMPS was at most $5.19\%$ lower than the performance of the optimal metalevel policy across all costs. The difference in performance was largest for the lowest cost $\lambda=0.001$ ($t(2999) = 3.75, p = 0.0002$) and decreased with increasing cost so that there was no statistically significant performance difference between BMPS and the optimal metalevel policy for costs greater than $\lambda=0.0025$ (all $p > 0.15$). BMPS performed between $6.78\%$ and $35.8\%$ better than the meta-greedy policy across all costs where the optimal policy made more than one observation (all $p < 0.0001$) and $20.3\%$ better on average ($t(44999)=42.4, p<10^{-15}$).

\subsection{4.2\space\space\space\space META-DECISION-MAKING}
How should an agent allocate its limited decision-time across estimating the expected utilities of multiple alternatives? To evaluate how well BMPS can solve this kind of problem, we evaluate it on the \emph{Bernoulli metalevel probability model} introduced by Hay et al. (2012). This problem is similar to the standard multi-armed bandit problem with one critical difference: Only the reward from the final pull counts---the previous "simulated" pulls provide information, but no reward. Like the first problem, the agent takes a single object-level action, choosing arm $i$ and receiving reward $r(s,a_i) \sim \text{Bernoulli}(\theta_i)$. Unlike the first problem, however, the agent must track multiple environment parameters and select among competing computations.

\subsubsection{4.2.1\space\space\space\space Metalevel MDP}
The Bernoulli metalevel probability model is a metalevel MDP $M\meta = (\B, \A, T\meta , r\meta, h)$ where each belief state $b$ defines $k$ Beta distributions over the reward probabilities $\theta_1,\cdots,\theta_k$ of the $k$ possible actions. Thus $b$ can be represented by $((\alpha_1, \beta_1), \dots , (\alpha_k, \beta_k))$ where $b(\theta_i) =  \text{Beta}(\theta_i; \alpha_i, \beta_i)$.
For the initial belief state $b_0$, these parameters are $\alpha_i = \beta_i = 1$. The metalevel actions $\A$ are $\{c_1, \dots , c_k, \bot\}$ where $c_i$ simulates action $a_i$ and $\bot$ terminates deliberation and executes the action with the highest expected return.
The metalevel transition function $T\meta$ encodes that performing computation $c_i$ increments $\alpha_i$ with probability $\frac{\alpha_i}{\alpha_i + \beta_i}$ and increments $\beta_i$ with probability $\frac{\beta_i}{\alpha_i + \beta_i}$. The metalevel reward function $r\meta(b, c)$ is $-\lambda$ for $c \in \left\lbrace c_1, \cdots, c_k \right\rbrace$ and $r\meta(b, \bot) = \max_i \frac{ \alpha_i }{ \alpha_i+\beta_i}$. Finally, the horizon $h$ is the maximum number of metalevel actions that can be performed and the last metalevel action must be $\bot$.

\subsubsection{4.2.2\space\space\space\space Evaluation procedure}
We evaluated BMPS on Bernoulli metalevel probability problems with $k \in \left\lbrace 2,\cdots,5 \right\rbrace$ object-level actions, a horizon of $h=25$, and computational costs ranging from $10^{-4}$ to $10^{-1}$. We compared the policy learned by BMPS with the optimal metalevel policy and three alternative approximations: the meta-greedy heuristic \cite{Russell1991}, the blinkered approximation \cite{Hay2012}, and the metalevel policy that always deliberates as much as possible. In addition to these, we also trained a Deep-Q-Network (DQN) \cite{Mnih2015} on the metalevel MDP to compare the performance of our method to baselines achieved by off-the-shelf deep RL methods \cite{baselines}.

We trained BMPS as described above, but with $10$ iterations of $1000$ episodes each.
To combat the possibility of overfitting, we evaluated the average returns of the five best weight vectors over $5000$ more episodes and selected the one that performed best. 
The relevance function for $\VPIsub$ matches each computation with the single parameter it is informative about, i.e., $f(c_j, i) = \mathds{1}(j = i)$.
The optimal metalevel policy and the blinkered policy were computed using backward induction \cite{Puterman2014}. The DQN was trained for $5,000,000$ steps. Since the episodes have a horizon of $h=25$, this resulted in more than $200,000$ training episodes for the DQN.
We evaluated the performance of each policy by its average return across $2000$ test episodes for each combination of computational cost and number of object-level actions.

\subsubsection{4.2.3\space\space\space\space Results}


We found that the BMPS policy attained 99.1\% of optimal performance ($0.6535$ vs. $0.6596$, $t(1998)= -7.43, p < 0.0001$) and significantly outperformed the meta-greedy heuristic ($0.60$, $t(1998)= 83.9, p <10^{-15}$), the full-deliberation policy ($0.20$, $t(1998)= 469.1, p <10^{-15}$), and the DQN ($0.58$, $t(1998)= 79.2, p <10^{-15}$). The performance of BMPS ($0.6535$) and the blinkered approximation ($0.6559$) differed by only $0.37\%$. 

\begin{figure*}[t!]
\centering
 \begin{subfigure}{0.45\textwidth}
   \includegraphics[width=\textwidth]{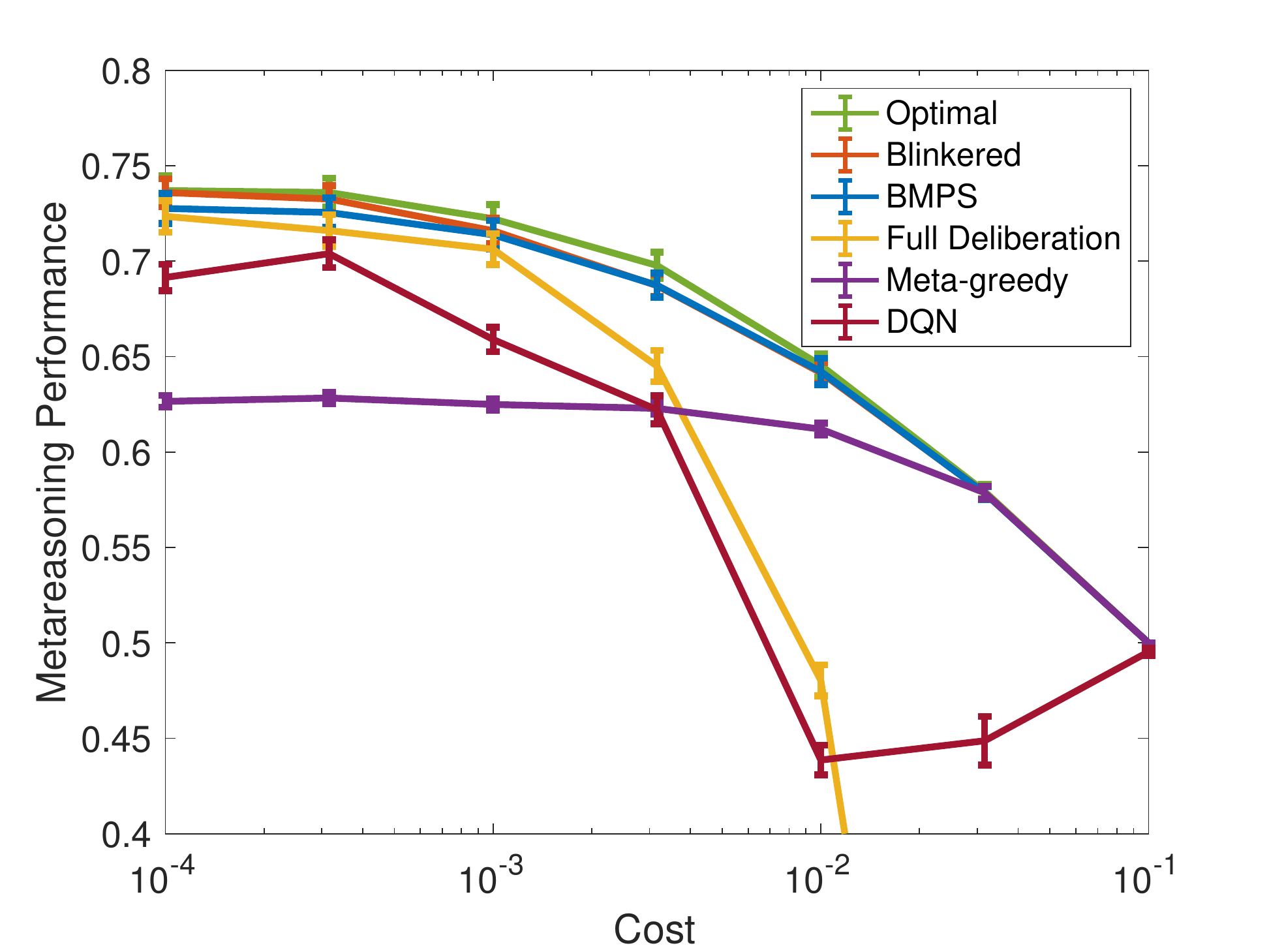}
   \caption{}
   \label{fig:MetbanditsCost}
\label{MetbanditsCost}
\end{subfigure}
\label{fig:MetaBandits}
\centering
\begin{subfigure}{0.45\textwidth}
   \includegraphics[width=\textwidth]{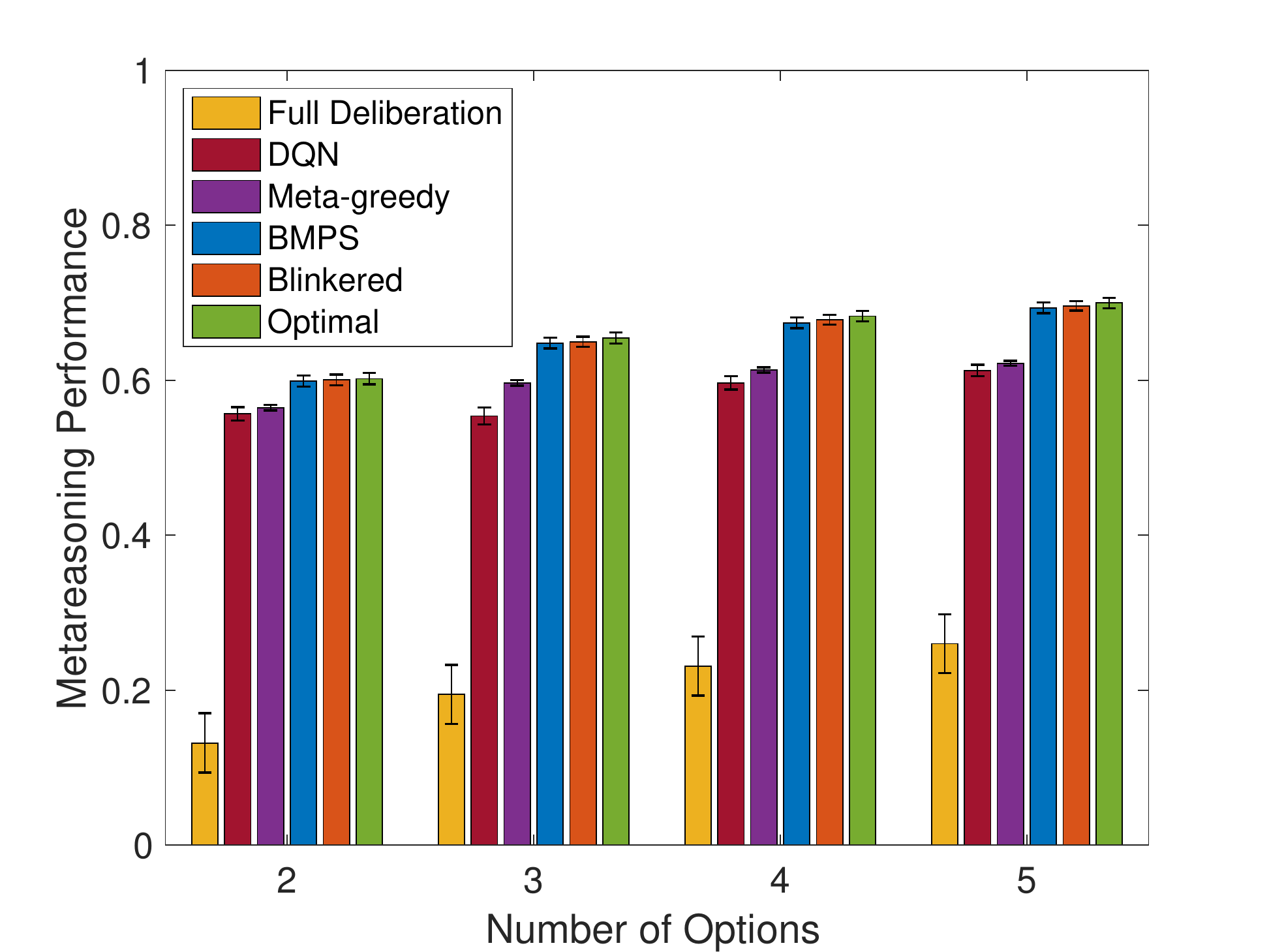}
   \caption{}
   \label{fig:MetbanditsArms}
\label{MetbanditsArms}
\end{subfigure}

\caption{
\label{Metabandits}
Metareasoning performance of alternative methods on the Bernoulli metalevel probability model (a) as a function of the cost of computation and (b) as a function of the number of actions. Metareasoning performance is defined as the expected reward for the chosen option minus the computational cost of the decision process. Error bars enclose $95\%$ confidence intervals.
}
 \end{figure*}

Figure \ref{fig:MetbanditsCost} shows the methods' average performance as a function of the cost of computation.
BMPS outperformed the meta-greedy heuristic for costs smaller than $0.03$ (all $p < 10^{-15}$), the full-deliberation policy for costs greater than $0.0003$ (all $p < 0.005$), and the DQN for all costs (all $p < 10^{-15}$). For costs below $0.0003$, the blinkered policy performed slightly better than BMPS (all $p < 0.01$). For all other costs both methods performed at the same level (all $p > 0.1$).
For costs above $0.01$, performance of BMPS  becomes indistinguishable from the optimal policy's performance (all $p>0.1$).

Figure \ref{fig:MetbanditsArms} shows the metareasoning performance of each method as a function of the number of options. 
We found that the performance of BMPS scaled well with the size of the decision problem. For each number of options, the relative performance of the different methods was consistent with the results reported above. 

Finally, as illustrated in Supplementary Figure 2, we found that BMPS learned surprisingly quickly, usually discovering near-optimal policies in less than 10 iterations. In particular, BMPS was able to perform significantly better than the DQN, despite being trained on fewer than $20\%$ as many episodes. This demonstrates the value of the proposed VOI features, which dramatically constrain the space of possible metalevel policies to be considered.

\subsection{4.3\space\space\space\space METAREASONING ABOUT PLANNING}

Having evaluated BMPS on problems of metareasoning about how to make a one-shot decision, we now evaluate its performance at deciding how to plan.
To do so, we define the \textit{Bernoulli metalevel tree}, which generalizes the Bernoulli metalevel probability model by replacing the one-shot decision between $k$ options by a tree-structured sequential decision problem that we will refer to as the \textit{object-level MDP}. The transitions of the object-level MDP are deterministic and known to the agent. The reward associated with each of $k = 2^{h+1}-1$ states in the tree is deterministic, but initially unknown; $r(s,a,s_i)=\theta_i \in \{-1, 1\}$. The agent can uncover these rewards through reasoning at a cost of $-\lambda$ per reward. When the agent terminates deliberation, it executes a policy with maximal expected utilty. Unlike in the previous domains, this policy entails a sequence of actions rather than a single action.

\subsubsection{4.3.1\space\space\space\space Metalevel MDP}
The Bernoulli metalevel tree is a metalevel MDP  $M\meta = (\B, \A, T\meta, r\meta)$ where each belief state $b$ encodes one Bernoulli distribution for each transition's reward. Thus, $b$ can be represented as $(p_1, \cdots, p_i)$ such that $b(\theta_i=1)=p_i$ and $b(\theta_i=-1)=1-p_i$.
The initial belief $b_0$ has $p_i = 0.5$ for all $i$.
The metalevel actions are defined $\A = \{c_1, \cdots, c_k, \bot\}$ where $c_i$ reveals the reward at state $s_i$ and $\bot$ selects the path with highest expected sum of rewards according to the current belief state.
The transition function $T\meta$ encodes that performing computation $c_i$ sets $p_i$ to $1$ or $0$ with equal probability (unless $p_i$ has already been updated, in which case $c_i$ has no effect).
The metalevel reward function is defined $r\meta(b, c) = -\lambda$ for $c \in \{c_1,\cdots, c_k \}$, and $r\meta(b, \bot) = \max_{\mathbf{t}\in \mathcal{T}} \sum_{i \in \mathbf{t}} \mathds{E}[ \theta_i \mid p_i ]$ where $\mathcal{T}$ is the set of possible trajectories $\mathbf{t}$ through the environment, and $\mathds{E}[ \theta_i \mid p_i ] = 2p_i - 1$ is the expected reward attained at  state $s_i$.

\subsubsection{4.3.3\space\space\space\space Evaluation procedure}
We evaluated each method's performance by its average return over $5000$ episodes for each combination of tree-height $h \in \{2, \cdots, 6\}$ and computational cost $\lambda \in \{ 2^{-7}, \cdots, 2^0 \}$. To facilitate comparisons across planning problems with different numbers of steps, we measured the performance of metalevel policies by their expected return divided by the tree-height.

We trained the BMPS policy with $100$ iterations of $1000$ episodes each. To combat the possibility of overfitting, we evaluated the average returns of the three best weight vectors over $2000$ more episodes and selected the one that performed best.
The relevance function for $\VPIsub$ maps a computation to all the parameters that affect the value of any policy that the initial computation is informative about, i.e. $f(c_j, i) = \mathds{1}(i \in \{j\} \cup \text{descendents}(j) \cup \text{ancestors}(j))$

For metareasoning about how to plan in trees of height 2 and 3, we were able to compute the optimal metalevel policy using dynamic programming. But for larger trees, computing the optimal metalevel policy would have taken significantly longer than $6$ hours and was therefore not undertaken.

The blinkered policy of Hay et al. (2012) is not directly applicable to planning because of its assumption of ``independent actions'' which is violated in the Bernoulli metalevel tree. Briefly, the assumption is violated because the reward at a given state affects the value of multiple policies. Thus, we derived a recursive generalization of the blinkered policy to compare with our method. See the Supporting Materials for details.

 \begin{figure*}[t!]
 \centering
 \begin{subfigure}{.45\textwidth}
  \centering
   \includegraphics[width=\textwidth]{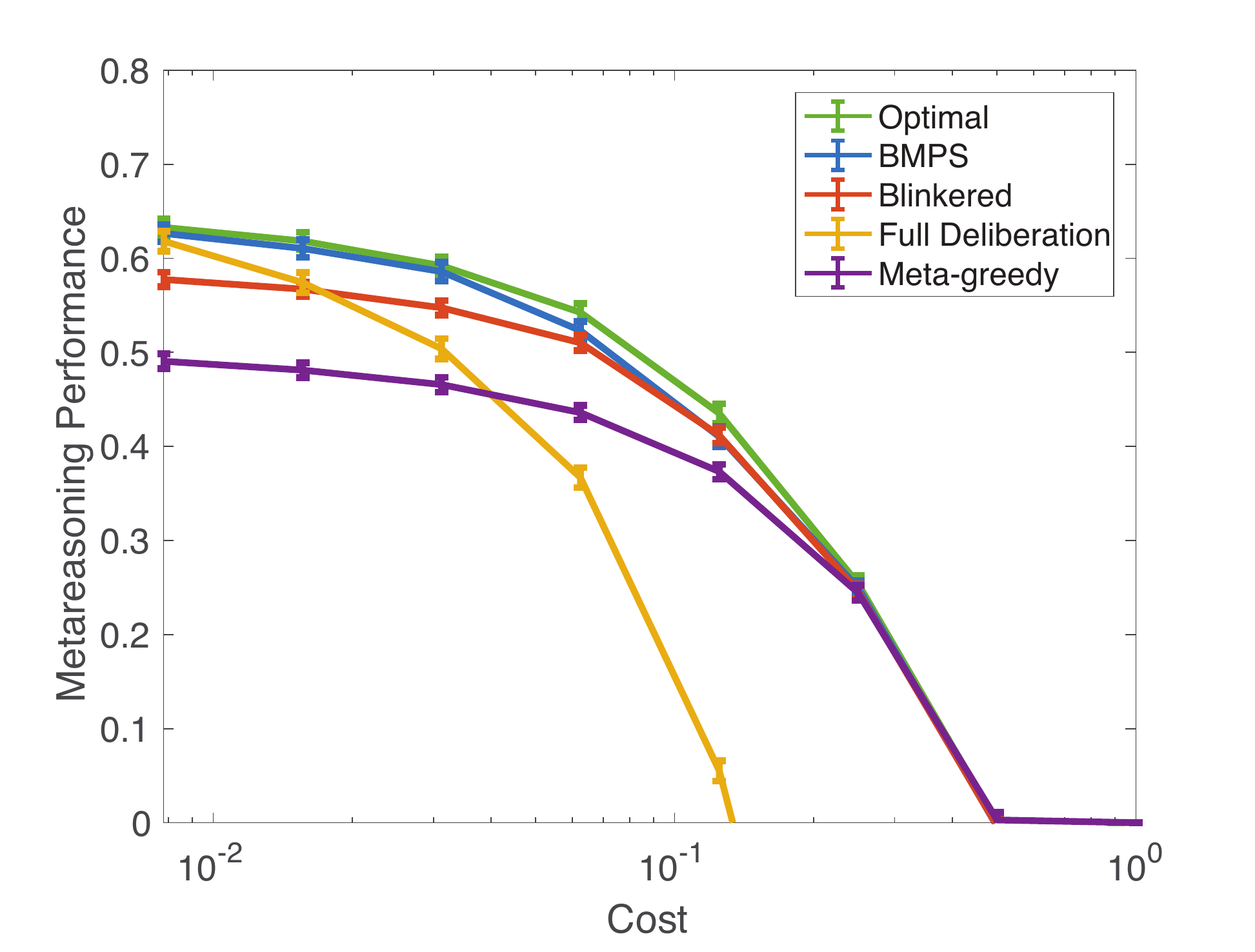}
  \caption{}
   \label{fig:BinaryTreesCost}
 \end{subfigure}
 \begin{subfigure}{.45\textwidth}
  \centering
   \includegraphics[width=\textwidth]{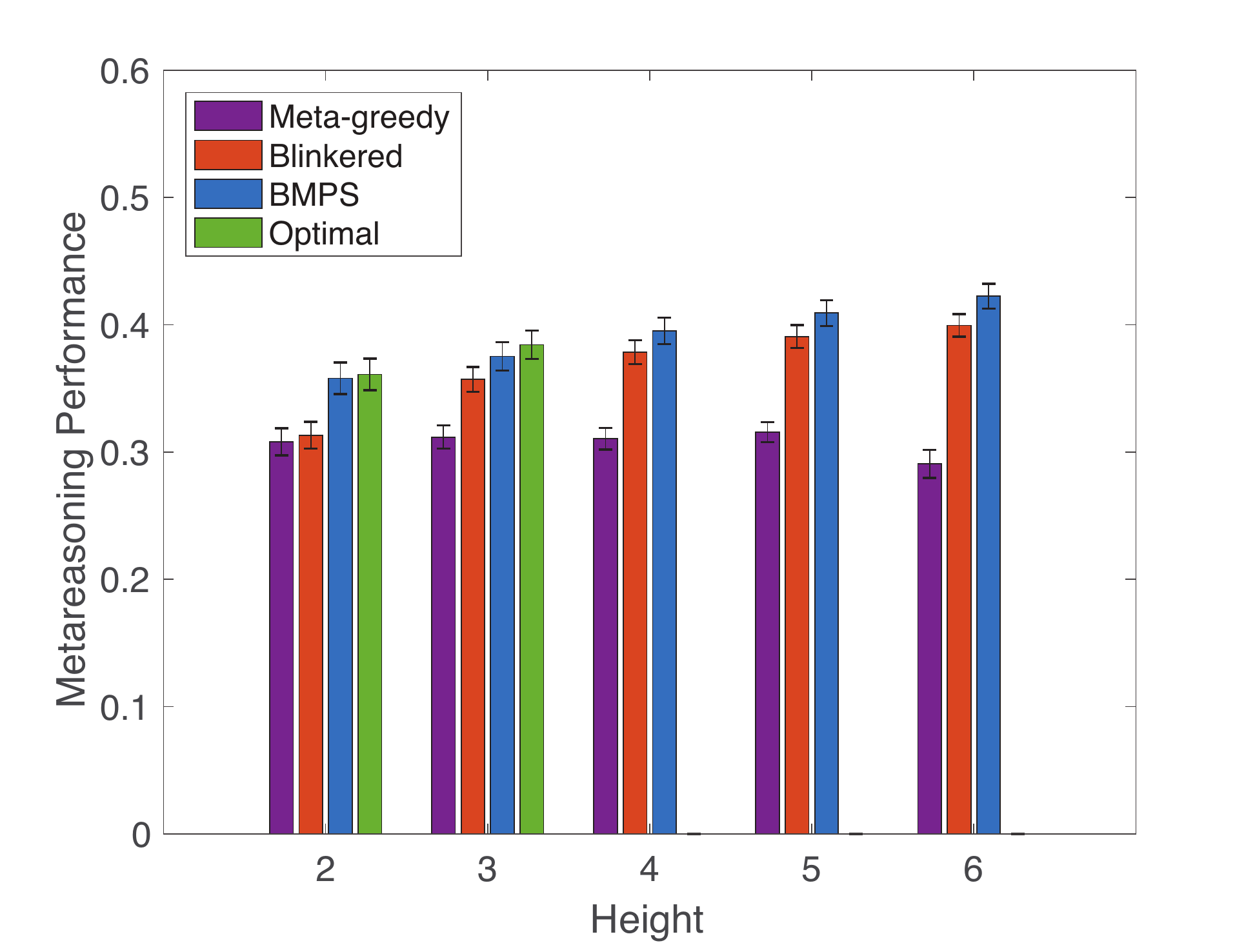}
  \caption{}
   \label{fig:BinaryTreesDepth}
 \end{subfigure}
\caption{
Metareasoning performance of alternative methods on the Bernoulli tree (a) as a function of computational cost (with tree-height 3) and (b) as a function of the number of actions (marginalizing over computational costs between $10^{-4}$ and $10^{-1}$). Metareasoning performance is normalized by tree height to facilitate comparison. In (b), the optimal policy is only shown for heights at which it can be computed in under six hours and the full observation policy is not shown because its performance is negative for all heights. Error bars enclose $95\%$ confidence intervals.
}
\label{fig:BinaryTrees}
\end{figure*}

\subsubsection{4.3.4\space\space\space\space Results}
We first compared BMPS with the optimal policy for $h \in \{2, 3\}$, finding that it attained 98.4\% of optimal performance ($0.367$ vs. $0.373$,  $t(159998) = -2.87$, $p<10^{-15}$).
%
Metareasoning performance differed significantly across the four methods we evaluated ($F(3, 799840) = 4625010; p < 10^{-15}$), and the magnitude of this effect depends on the height of the tree ($F(12,799840) = 1110179, p < 10^{-15}$) and the cost of computation ($F(21,799840)=1266582, p < 10^{-15}$).

Across all heights and costs, BMPS achieved a metareasoning performance of 0.392 units of reward per object-level action, thereby outperforming the meta-greedy heuristic ($0.307,\ t(399998) = 72.84,\ p < 10^{-15}$), the recursively blinkered policy ($0.368,\ t(399998) = 20.77,\ p < 10^{-15}$), and the full-deliberation policy ($-1.740,\ t(399998) = 231.18,\ p < 10^{-15}$).

As shown in Figure \ref{fig:BinaryTrees}a, BMPS performed near-optimally across all computational costs, and its advantage over the meta-greedy heuristic and the tree-blinkered approximation was largest when the cost of computation was low, whereas its benefit over the full-deliberation policy increased with the cost of computation.

Figure \ref{fig:BinaryTrees}b shows that the performance of BMPS scaled very well with the size of the planning problem, and that its advantage over the meta-greedy heuristic increased with the height of the tree. 

\section{5\space\space\space\space IS METAREASONING USEFUL?}
The costs of metareasoning often outweigh the resulting improvements in object-level reasoning. But here we show that the benefits of BMPS outweigh its costs in a potential application to emergency management.

During severe weather, important decisions---such as which cities to evacuate in the face of an approaching tornado---must be based on a limited number of computationally intense weather simulations that estimate the probability that a city will be severely hit \cite{baumgart2008emergency}.
Based on these simulations, an emergency manager makes evacuation decisions so as to minimize the risk of false positive errors (evacuating cities that are safe) and false negative errors (failing to evacuate a city the tornado  hits). We assume that the manager has access to a single supercomputer, but pays no cost for running each simulation. Thus, the manager has a fixed budget of simulations and her goal is to maximize the expected utility of the final decision.

\subsection{5.1\space\space\space\space METHODS}
We model the above scenario as follows: There is a finite amount of time $T$ until evacuation decisions about $k$ cities have to be made. For each city $i$, the emergency manager can run a fine grained, stochastic simulation ($c_i$) of how it will be impacted by the approaching tornado. Each simulation yields a binary outcome, indicating whether the simulated impact would warrant an evacuation or not. The belief state $b$ and transition function $T\meta$ of the corresponding metalevel MDP are the same as in the Bernoulli metalevel probability model: Each belief state defines $k$ Beta distributions that track the probability that the tornado will cause evacuation-warranting damage in each city. The parameters $\alpha_i$ and $\beta_i$ correspond to the number of simulations predicting that the tornado \{would | would not\} be strong enough to warrant an evacuation of city $i$. Prior to the first simulation, the parameters for each city $i$ are initialized as $\alpha_i=0.1$ and $\beta_i=0.9$ to capture the prior knowledge that evacuations are rarely necessary.
The primary formal difference from the Bernoulli metalevel probability model lies in how the final belief state is translated into a decision and reward. Rather than choosing a single option, the agent must make $k$ independent binary decisions about whether to evacuate each city. Evacuation has a cost, $\lambda_{\text{evac}}=-1$, but failing to evacuate a heavily-hit city has a much larger cost,  $\lambda_{\text{fn}}=-20$. Thus, the metalevel reward function is
\begin{equation}
    r\meta(b,\bot) = \sum_{1 \leq i \leq k}  \max \left\lbrace \frac{\alpha_i}{\alpha_i+\beta_i}\cdot \lambda_{\text{fn}},  \lambda_{\text{evac}} \right\rbrace.
\end{equation}

In contrast to the previous simulations, we now explicitly consider the cost of metareasoning. The decision time $T$ has to be allocated between reasoning about the cities and metareasoning about which city to reason about so that
$T = n_{\text{sim}} \cdot ( t_{\text{MR}} + t_{\text{sim}} )$,
where $n_{\text{sim}}$ is the number of simulations run, $t_{\text{MR}}$ is the amount of time it takes to choose one simulation to run (i.e. by metareasoning), and $t_{\text{sim}}$ is the amount of time it takes to run one simulation. Thus, for given values of $t_{\text{MR}}$ and $t_{\text{sim}}$ the number of simulations that can be performed is 
$
    n_{\text{sim}} = \left\lfloor \frac{T}{t_{\text{MR}} + t_{\text{sim}}} \right\rfloor,
$
where $\lfloor x \rfloor$ rounds $x$ down to the closest integer. Note that metalevel policy is computed offline, and thus training time does not factor into the above equation.  The simulations reported below use a single BMPS policy optimized for $k=20$ and $n_{\text{sim}}=50$ to mimic the reuse of pre-computed weights in practical applications; the weights are relatively insensitive to these parameters.




To assess if BMPS could be useful in practice, we compare the utility of evacuation decisions made by its metalevel policy to those made by a baseline metalevel policy that uniformly distributes simulations across the $k$ cities. Since the BMPS policy has $t_{\text{MR}} > 0$ while the baseline policy has $t_{\text{MR}} \approx 0$, BMPS will typically run fewer simulations 
and must make up for this by choosing more valuable ones.


\subsection{5.2\space\space\space\space RESULTS}
We evaluated the BMPS policy and the uniform computation policy on the tornado problem with $T=24$ hours, $k \in \{10, 30\}$ cities, and a range of plausible values for the duration of each weather simulation ($t_{\text{sim}} \in [2^{-2}, 2^{4}]$ hours). For each policy and parameter setting we estimate utility as the mean return over $5000$ rollouts.

\begin{figure}[b!]
    \centering
    \includegraphics[width=0.45\textwidth]{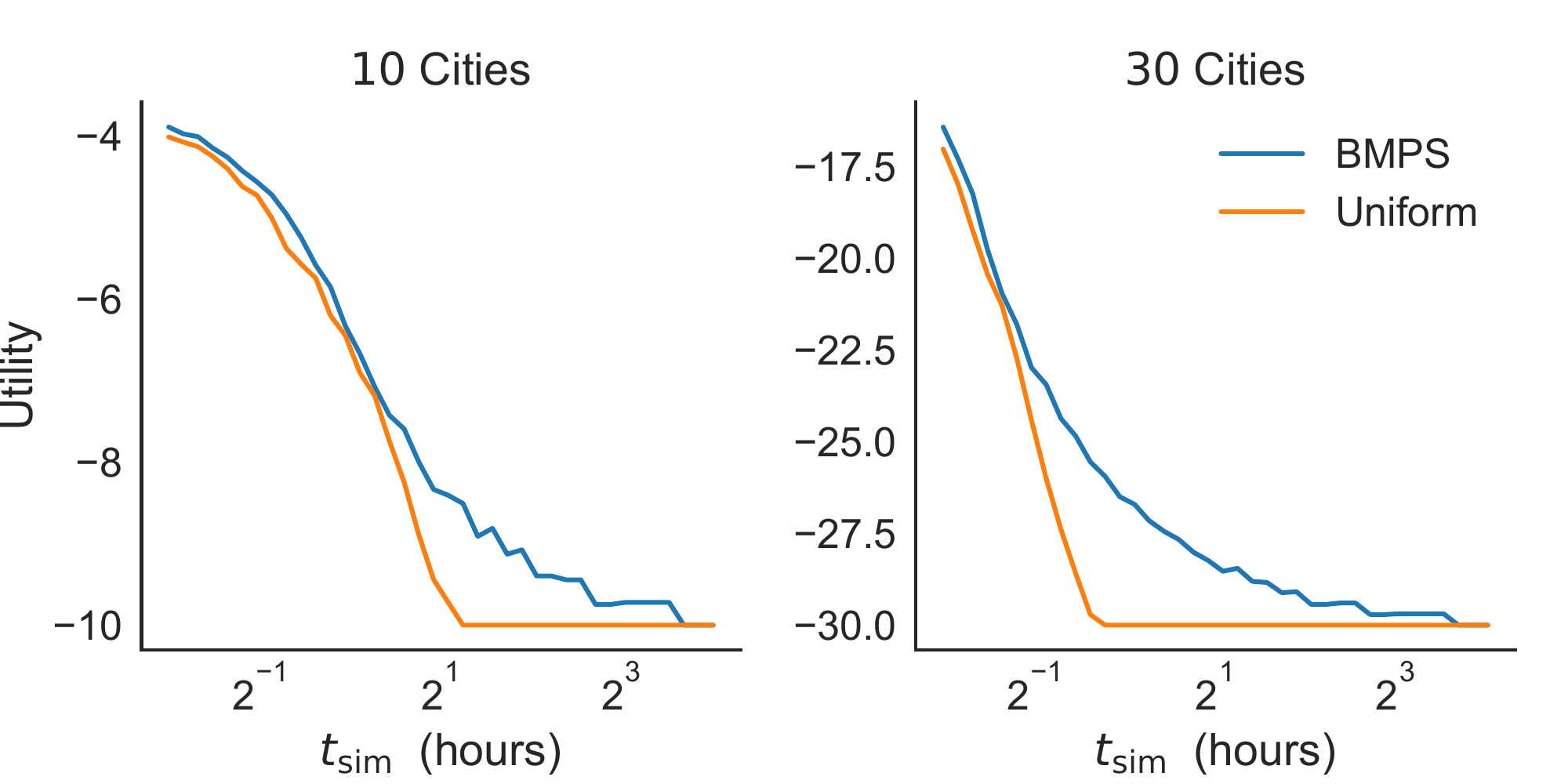}
    \caption{Benefit of metareasoning in the tornado evacuation scenario depending on the duration of each simulation ($t_{\text{sim}}$) and the number of cities considered.}
    \label{fig:tornado-sims}
 \end{figure}

Empirically, we found that $t_{\text{MR}} \approx 1~\text{ms}$ for $k=10$ and $t_{\text{MR}} \approx 3~\text{ms}$ for $k=30$. Thus, even with a conservative estimate of $t_{\text{MR}}=0.001$ hours, metareasoning would cost at most one simulation. Consequently, in our simulations, diverting some of the computational resources to metareasoning was advantageous regardless of how long exactly a tornado simulation might take and the number of cities being considered. As Figure \ref{fig:tornado-sims} shows, the benefit of metareasoning was larger for the more complex problem with more cities and peaked for an intermediate cost of object-level reasoning.


While this is a hypothetical scenario, it suggests that BMPS could be useful for practical applications. Specifically, we suggest that the method will be most valuable when a metareasoning problem must be faced multiple times (so that the cost of training BMPS offline can be amortized) and object-level computations are expensive (so that the resulting savings in object-level reasoning outweigh the online cost of computing the features used for metareasoning). In follow-up simulations, we explored conditions in which the cost of metareasoning causes a substantial reduction in the number of simulations that can be run. We found that metareasoning continues to be useful as long as object-level computation is substantially more expensive than metareasoning (see Supplementary Material).



\section{6\space\space\space\space DISCUSSION}

We have introduced a new approach to solving the foundational problem of rational metareasoning: metacognitive reinforcement learning. This approach applies algorithms from RL to metalevel MDPs to learn a policy for selecting computations. 
Our results show that BMPS can outperform the state of the art for approximate metareasoning. While we illustrated this approach using a policy search algorithm based on Bayesian optimization, there are many other RL algorithms that could be used instead, including policy gradient algorithms, actor-critic methods, and temporal difference learning with function approximation.

Since BMPS approximates the value of computation as a mixture of the myopic VOI and two other VOI features, it can be seen as a generalization of the meta-greedy approximation \cite{Lin2015,RussellWefald1991}. It is the combination of these features with RL that makes BMPS tractable and powerful.
BMPS works well across a wider range of problems than previous approximations because it reduces arbitrarily complex metalevel MDPs to low-dimensional optimization problems.
We predict that metacognitive RL will enable significant advances in artificial intelligence and its applications. In the long view, metacognitive RL may become a foundation for self-improving AI systems that learn how to solve increasingly complex problems with increasing efficiency. 

One weakness of our approach is that the time required to compute the value of perfect information by exact integration increases exponentially with the number of parameters in the agent's model of the environment. Thus, an important direction for future work is developing efficient approximations or alternatives to this feature, and/or discovering new features via deep RL \cite{Mnih2015}. A second limitation is our assumption that the meta-reasoner has an exact model of its own computational architecture in the form of a metalevel MDP. This motivates the incorporation of model-learning mechanisms into a metacognitive RL algorithm.


We have shown that the benefits of metareasoning with our method already more than outweigh its computational costs in scenarios where the object-level computations are very expensive. It might therefore benefit practical applications that involve complex large-scale simulations, active learning problems, hyperparameter search, and the optimization of functions that are very expensive to evaluate. Finally, BMPS could also be applied to derive rational process models of human cognition. 

\pagebreak

\bibliographystyle{apacite}
{\footnotesize
\bibliography{references}
}

\pagebreak

\section{SUPPLEMENTARY MATERIAL}

\subsection{THE RECURSIVELY BLINKERED POLICY}
The blinkered policy of Hay et al. (2012) was defined for problems where each computation informs the value of only one action. This assumption of ``independent actions'' is crucial to the efficiency of the blinkered approximation because it allows the problem to be decomposed into independent (and easily solved) subproblems for each action. However, the assumption does not hold for the Bernoulli metalevel tree because the reward at a given state affects the value of multiple policies. This is because in the context of sequential decision making, ``actions'' become policies, and the reward at one state affects the values of all policies visiting that state. Thus, a single computation affects the value of many policies. An intuitive generalization would be to approximate the value of a computation $c$ by assuming that future computations will be limited to those that are informative about \textit{any} of the policies the initial computation is relevant to, a set we call $\mathcal{E}_{c}$,. However, for large trees, this only modestly reduces the size of the initial problem. This suggests a recursive generalization: Rather than applying the blinkered approximation once and solving the resulting subproblem exactly, we recursively apply the approximation to the resulting subproblems. Finally, to ensure that the subproblems decrease in size monotonically, we remove from $\mathcal{E}_{c}$ the computations about rewards on the path from the agent's current state to the state inspected by computation $c$ and call the resulting set $\mathcal{E}_{c}^\prime$. Thus, we define the \emph{recursively blinkered policy} as $\pi^{\text{RB}}(b) = \arg\max_c Q^{\text{RB}}(b, c)$ with $Q^{\text{RB}}(b_t, \bot) = r\meta(b_t, \bot)$ and $Q\meta^{\text{RB}}(b, c) =$
\begin{equation*}
r\meta(b, c) + \expect[B'\sim T\meta(b,c,\cdot)]{
    \max_{c' \in \mathcal{E}_{c'}^\prime}
        Q^{\text{RB}}(B', c')
}
\end{equation*}

\subsection{DETAILS ON SIMULATIONS REPORTED IN SECTION 5}
We found the computational cost of metareasoning for the tornado problem to be several orders of magnitude lower than realistic costs of \textit{object}-level computations (i.e. weather simulations). Thus, the simulations leave open the question of whether BMPS can also be usefully applied when metareasoning costs are non-negligible. To answer this question, we ran additional simulations for the tornado problem with \textit{un}realistically low values of $T$ and $t_{\text{sim}}$.

\begin{figure}[h!]
    \centering
    \includegraphics[width=0.45\textwidth]{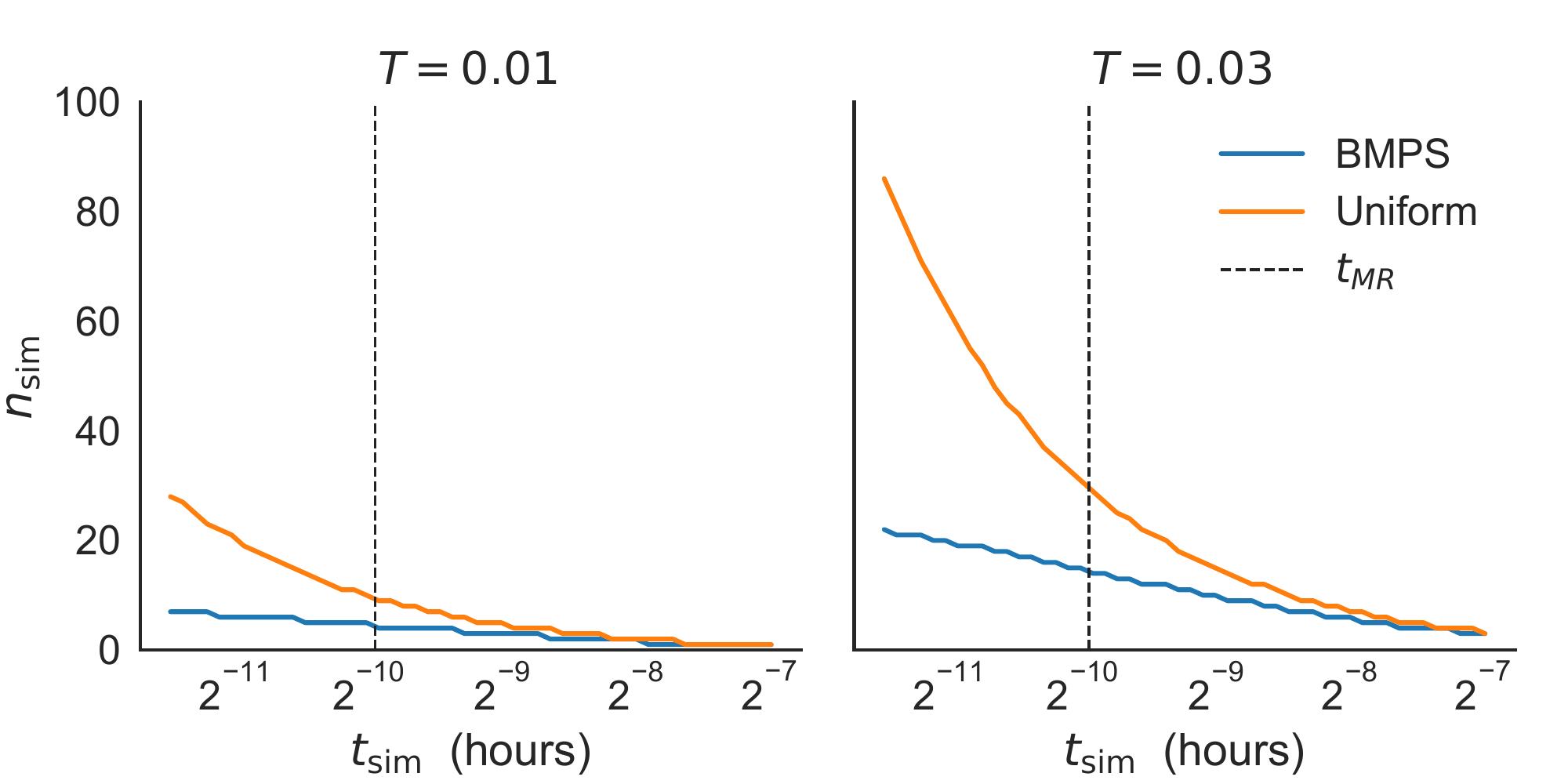}
    \caption{The number of simulations that can be run with versus without metareasoning as a function of the total time $T$ and the cost of each simulation $t_{\text{sim}}$.}
    \label{fig:abstract-nsim}
 \end{figure}

\begin{figure}[h!]
    \centering
    \includegraphics[width=0.45\textwidth]{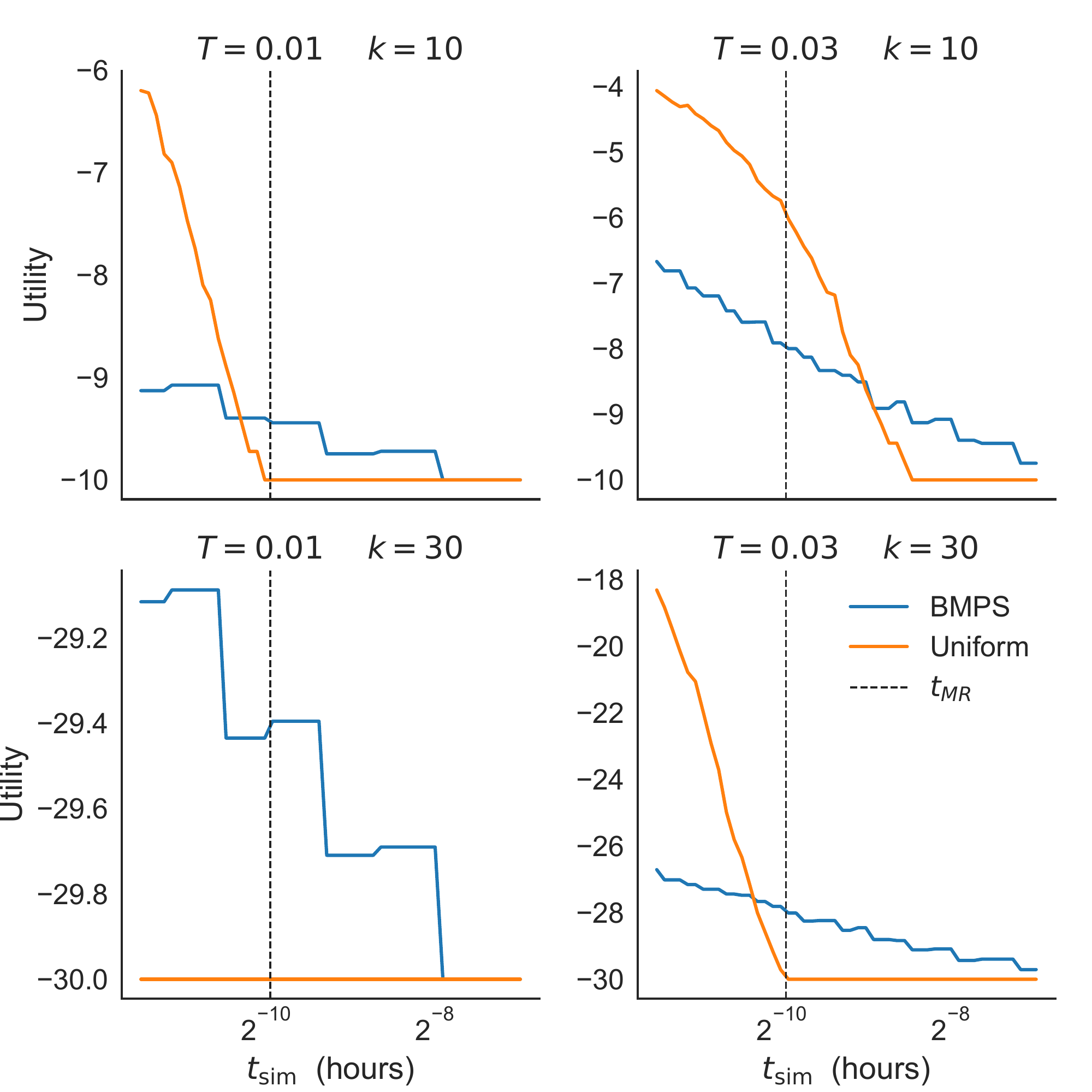}

    \caption{Utility of BMPS vs. allocating computation uniformly as a function of the total time $T$, the cost of each simulation $t_{\text{sim}}$, and the number of possible computations (i.e. the number of cities) $k$.}
    \label{fig:abstract}
 \end{figure}

The simulations summarized in Figure~\ref{fig:abstract-nsim} investigated hypothetical scenarios where the metareasoning cost incurred by the BMPS policy considerably reduces the amount of object-level computation it can perform. This reduction is greatest when object-level computations are fast and the total amount of available time $T$ is high. Nevertheless, as shown in Figure \ref{fig:abstract}, BMPS still often outperforms allocating computation time uniformly. This is often true even when BMPS can perform only half as many simulations (e.g. $T = 0.03;\ k=30;\ t_{\text{sim}}=2^{-10}$). As expected, when the time to run a simulation is much less than the time to metareason about which simulation to run, metareasoning does not pay off anymore. Overall, we see that the benefit of metareasoning increases with the costliness of object-level reasoning and the number of computations that must be considered, but decreases with increased total computation time.


\renewcommand{\arraystretch}{1.4}
\newcolumntype{b}{>{\hsize=1.7\hsize}X}
\newcolumntype{s}{>{\hsize=.3\hsize}X}
\begin{table*}[hb!]
    \centering
    \begin{tabular*}{.8\textwidth}{p{.15\textwidth} p{.6\textwidth}}
        \toprule
        $\mathcal{M}\meta$ & meta-level Markov Decision Process \\
        $\B$ & Set of possible belief states \\ 
        $\mathcal{A}$ &  Set of meta-level actions $\C, \cup \{ \bot \} $\\
        $\C$ & set of possible computations\\ 
        $\bot$ & meta-level action that terminates deliberation and initiates an object-level action\\
        $r\meta(b,c)$  & reward function of the meta-level MDP, $r\meta(b,c)=-\text{cost}(c)=-\lambda$ for $c \in \C$ and $r\meta(b,\bot) = \max_\pi \Etheta{\uthetapi}$ \\
        $\lambda$ & cost of a single computation \\
        $T\meta(b,c,b')$ & probability that performing computation $c$ in belief state $b$ leads to belief state $b^\prime$ \\
        $\theta $ & parameters of the agent's model of the environment \\  
        $\pi$ & object-level policy for selecting physical actions \\ 
        $\uthetapi$ & expected return of acting according to the object-level policy $\pi$ if $\theta$ is the correct model of the environment\\ 
        $U(b)$ & expected value of terminating computation with the belief $b$, $r\meta(b, \bot)$ \\
        $\pi\meta$ & meta-level policy for selecting computational actions\\ 
        $\pi\meta^\star$ & optimal meta-level policy, see Equation~\ref{eq:optimalPolicy} \\
        $\text{VOC}(c,b)$ & Value of Computation, the expected improvement in decision quality that can be achieved by performing computation $c$ in belief state $b$ and continuing optimally, minus the cost of the optimal sequence of computations \\
        $\text{VOI}_1(c,b)$ & myopic Value of Information, expected improvement in decision quality from taking a single computation $c$ before terminating computation, see Equation \ref{eq:VOI1} \\
        $\VPIall(b)$ & Value of Perfect Information, the expected improvement in decision quality from attaining a maximally informed belief state beginning in belief state $b$,
        see Equation~\ref{eq:VPIall} \\
        $\VPIsub(c,b)$ & value of attaining perfect information about the subset of components of $\theta$ that are most relevant to computation $c$, see Equation~\ref{eq:VPIsub}\\
        \bottomrule
\end{tabular*}
\caption{Mathematical notation}
\label{tab:Notation}
\end{table*}

\end{document}